\documentclass[runningheads]{llncs}

\usepackage[T1]{fontenc}
\usepackage{graphicx}
\usepackage{algorithm}
\usepackage{algpseudocodex}   
\usepackage{amsmath,bm}
\usepackage{amsfonts}
\usepackage{float}
\usepackage{hyperref}
\usepackage{booktabs}
\usepackage{fontawesome5}

\begin{document}
\title{Conditional Flow Matching for Cross-Field MRI Harmonisation}
\author{Baris Imre \inst{1} \href{mailto:b.imre@lumc.nl}{\faEnvelope}, Aram Salehi\inst{2,3}, Levente Baljer\inst{4}, Andrew Webb\inst{3, 5}, Marius Staring \inst{1}, Efe Ilicak \inst{1}}
\authorrunning{B. Imre et al.}

\institute{Division of Image Processing, Department of Radiology, Leiden University Medical Center \and Department of Human Genetics, Radboud University Medical Center \and Department of Radiology, Leiden University Medical Center \and Department of Neuroimaging, King’s College London \and Department of Bioengineering, University of Illinois}

\maketitle

\begin{abstract}
Magnetic resonance images of the same subject look markedly different across field strengths, which complicates the comparison and pooling of data across sites. We address cross-field brain-MRI translation for the MRIxFields2026 challenge, and in particular its Task~3: a \emph{single} model that translates between \emph{any} directed pair of the five field strengths and across three contrasts. We phrase the problem as a conditional flow matching path: because the source and target volumes are spatially registered, we learn a velocity field that carries the source slice directly to the target slice, rather than starting from noise. To learn this mapping from only three paired subjects, the unified model is trained in three stages: a degradation-bridge pretraining that distills a restoration prior from the abundant unpaired retrospective cohort, a cross-field finetuning over all directed pairs on the paired cohort, and an adversarial refinement that sharpens the output. At inference, we integrate the learned velocity with a second-order Heun solver in a handful of steps. A restoration prior learned without any paired data already reaches a mean SSIM of 0.837, and each subsequent training stage improves on it. A single 6.3M-parameter model thereby covers all 60 field-pair and contrast combinations, with inference in five solver steps per slice. On the challenge evaluation set the model reaches a mean SSIM of 0.909, averaged over the three contrasts, outperforming regression and diffusion baselines built on the identical network on all three challenge metrics.

\end{abstract}

\section{Introduction}
Magnetic resonance imaging is central to neuroimaging. Typical MRI examinations consist of weighted images, as opposed to quantitative maps, that depend on tissue properties as well as acquisition settings. This makes images hard to compare across sites and hardware, hindering the reproducibility and the pooling of data for large studies.

Field strength heavily contributes to the signal-to-noise ratio and the clinically achievable spatial resolution, and shifts tissue contrast. The same subject can look different at different fields, even for the same protocol. Translating a scan from one field to another would put these images on common ground.

The MRIxFields2026 challenge poses this as cross-field MRI translation across five field strengths and three contrasts. There are three tasks. Task~1 maps any field up to 7T, Task~2 maps 0.1T up to any higher field, and Task~3 demands a single model that translates between \emph{any} directed pair of fields. In this paper we mainly focus on Task~3.
We propose a conditional flow matching method that bridges the source slice to the target with a time-dependent velocity field, conditioned on the timestep and the source and target field strengths through FiLM, with the three contrasts stacked as input channels. The unified model is trained in three stages: a degradation-bridge pretraining that learns a restoration prior from unpaired data, a cross-field finetuning over all directed pairs, and an adversarial refinement that sharpens the output.
\section{Materials and Methods}
\begin{figure}[t]
    \centering
    \includegraphics[width=\textwidth]{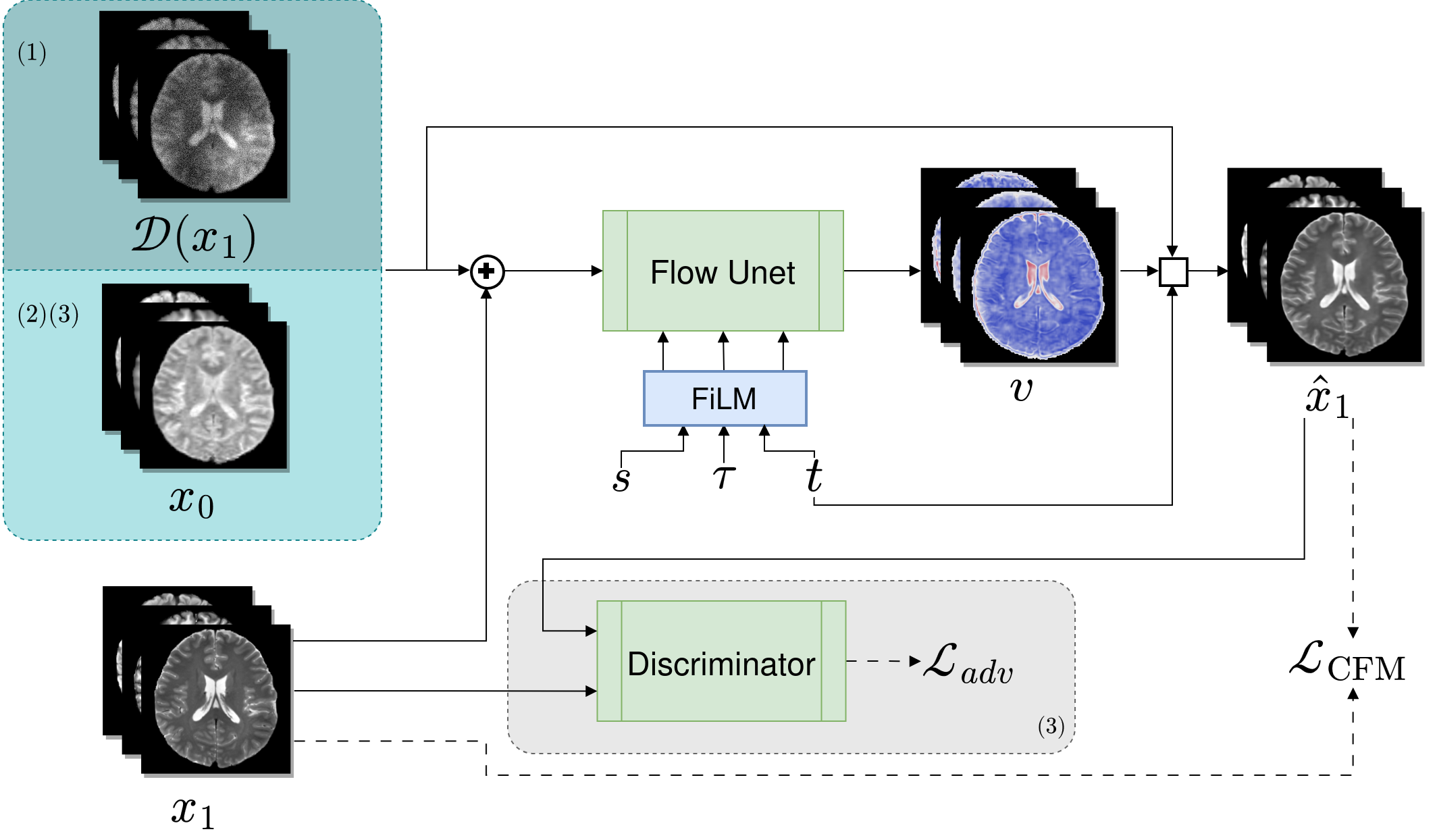}
    \caption{The overall training architecture. Training is split into three stages, shown here as $(1), (2), (3)$. In the pretraining stage the source ($x_0$) is a \emph{degraded copy of the target} ($x_1$), since this cohort is unpaired; the network learns to restore it. In the second and third stages we continue with paired training. In the third stage, adversarial refinement, a discriminator judges the outputs of our CFM model and supplies an adversarial loss. The source field $s$, target field $\tau$, and the timestep $t$ are embedded and injected through FiLM. During training the network predicts a velocity that transports the source towards the target; at inference this velocity is integrated over several steps.}
    \label{fig:overview}
\end{figure}
\subsection{Dataset}
We use the MRIxFields2026 multi-field brain-MRI dataset, which spans five field strengths (0.1T, 1.5T, 3T, 5T, 7T) acquired on scanners from three manufacturers, with three contrasts per field: T1w, T2w, and T2-FLAIR. The data comprise two cohorts. A retrospective, multi-centre cohort provides \emph{unpaired} volumes, in which each subject is scanned at a single field strength, and where some subjects are missing contrasts. A prospective \textit{travelling} cohort provides \emph{paired} volumes, in which the same subject is scanned at every field strength, yielding spatially registered source and target pairs. The travelling training set for this challenge comprises only three subjects, each scanned with all three contrasts at all five field strengths -- the central data constraint of our setting.
Task~3 must be solved by a \emph{single} set of parameters: with five fields there are $5\times4=20$ directed pairs, and with three contrasts $20\times3=60$ subtasks, and the rules forbid an ensemble or router of per-pair specialists. Submissions are scored on voxel metrics only, normalised RMSE (nRMSE), structural similarity (SSIM) \cite{wang2004ssim}, and LPIPS \cite{zhang2018lpips}, over the central axial slab (slices $[150,180)$) of each volume. The validation cohort ships inputs only, with cross-field ground truth withheld on the server, so predictions cannot be scored locally and are instead inspected qualitatively (Sect.~\ref{sec:results}).

\subsection{Conditional flow matching}

Conditional flow matching (CFM) \cite{flow_matching} learns to transport an initial distribution $\pi_0$ to a target $\pi_1$ along a smooth path, by regressing a time-dependent velocity field that carries samples $x_0 \sim \pi_0$ to $x_1 \sim \pi_1$. Like diffusion, it is trained by regression, but unlike the stochastic sampling of diffusion \cite{ddpm} it defines a deterministic transport and generates samples by integrating an ODE. The network $v_\theta(x, t)$ is regressed onto the velocity of a conditional path; we take the linear interpolation between paired endpoints $x_0 \sim \pi_0$ and $x_1 \sim \pi_1$,

\begin{equation}
    x_t = (1-t)\, x_0 + t\, x_1, \quad t \sim \mathcal{U}[0, 1].
\end{equation}
The velocity of this path is constant and given by its time derivative, $\dot{x}_t = x_1 - x_0$, which serves as the regression target \cite{rectified_flow,liu2023rectifiedflow}. The network is trained by minimising

\begin{equation}
    \mathcal{L}_{\mathrm{CFM}}(\theta)
    = \mathbb{E}_{t,\, x_0,\, x_1}
      \left\| v_\theta(x_t, t) - (x_1 - x_0) \right\|^2 .
\end{equation}
Because the paired volumes are spatially registered, we bridge directly between the source slice ($x_0$) and the target slice ($x_1$) instead of noise. The source enters only at the $t=0$ state of the integration, so the network sees the interpolated $x_t$ and never receives the source as a separate input, keeping the bridge leak-free.

\subsection{Neural network conditioning}

The three contrasts are co-registered per subject and stacked as input channels, so the network predicts a three-channel velocity in a single pass. Because one set of parameters must serve every directed pair, the network cannot infer the intended translation from the input alone and is told which fields it maps between through three signals: the timestep $t$, the source field $s$, and the target field $\tau$. The timestep uses a sinusoidal embedding and $s, \tau$ two learned embedding tables, summed into a single conditioning vector:
\begin{equation}
    c = \mathrm{emb}_t(t) + \mathrm{emb}_s(s) + \mathrm{emb}_\tau(\tau).
\end{equation}
The conditioning vector $c$ modulates the network through feature-wise linear modulation (FiLM) \cite{film}. Inside each residual block, a single linear map projects $c$ and splits its output into a per-channel scale $\gamma(c)$ and shift $\beta(c)$, which are applied to the group-normalised feature map $h$,
\begin{equation}
    \mathrm{FiLM}(h \mid c) = \big(1 + \gamma(c)\big) \odot \mathrm{norm}(h) + \beta(c),
\end{equation}
where $\odot$ broadcasts the per-channel scale over spatial dimensions. The scale weights are zero-initialised, so modulation starts at identity for a stable start.

\subsection{Inference}
At inference we translate a source slice by integrating the learned velocity field forward in time, from the source at $t=0$ to the target at $t=1$. The source slice is the initial condition $x_{t=0}=x_0$. We discretise $[0,1]$ into $N$ uniform steps and integrate with Heun's method, a second-order predictor--corrector scheme:
\begin{align}
    \tilde{x}_{t+h} &= x_t + h\, v_\theta(x_t, t,\, s,\, \tau), \\
    x_{t+h} &= x_t + \tfrac{h}{2}\left[ v_\theta(x_t, t,\, s,\, \tau) + v_\theta(\tilde{x}_{t+h}, t+h,\, s,\, \tau) \right],
\end{align}
where the Euler step $\tilde{x}_{t+h}$ is the predictor and the update averages the velocities at the current and predicted states. This tracks the trajectory more accurately than a single Euler step, so far fewer steps are needed and inference stays cheap; the same solver choice proved effective for diffusion ODEs by Karras et al. \cite{heun}.

\subsection{Adversarial refinement}

Regressing the velocity under a squared-error objective yields a mean estimator: wherever the mapping is uncertain, the network averages the plausible high-frequency detail, which appears as blur. We counter this with a generative adversarial \cite{gan} term, rewarding the generator (in our case: the flow matching model) for producing samples a discriminator $D$ cannot tell from real, which restores realistic texture the CFM output lacks.

We use a conditional, multi-scale PatchGAN discriminator \cite{patchgan}, which emits a grid of judgments over local receptive fields and so scores texture rather than global anatomy. It is conditioned on the same signals as the generator, the source $x_0$ and the fields $s, \tau$, so it judges plausibility for that specific translation. It is trained with a hinge adversarial loss, and stabilised by a feature-matching \cite{feature_matching} term on $D$'s intermediate activations. Applied only as a final stage on top of the trained CFM model, the refinement keeps a standard velocity-regression step alongside the adversarial and feature-matching terms; this \textit{faithfulness anchor} ties the output to the correct translation and stops the adversarial pressure from drifting off target.

\subsection{Staged training}
We train in three stages: degradation-bridge pretraining on the unpaired retrospective cohort, then cross-field finetuning, and adversarial refinement on the paired travelling cohort \cite{stages_flow}. The first stage builds a field conditioned prior from the abundant unpaired data; the latter two adapt that prior to the directed source-to-target mapping using the highly limited paired data.

The first stage, \emph{degradation-bridge pretraining} (Algorithm~\ref{alg:pretrain}), trains a restoration prior on the unpaired retrospective data. We bridge from a \emph{degraded copy of the target} to the clean target: for a real slice $x_1$ at field $\tau$ we synthesise a corrupted $x_0 = \mathcal{D}(x_1)$ and regress the conditional flow matching velocity that transports $x_0$ to $x_1$. The source field is left unspecified ($\emptyset$), and a per-channel mask $m$ confines the loss to the contrasts a given subject actually has.

The operator $\mathcal{D}$ is an array of MRI motivated corruptions, each drawn and applied independently per sample with probability $0.4$--$0.5$: additive noise ($\sigma \in [0.02, 0.3]$), Gaussian blur ($\sigma \in [0.5, 2.5]$\,px), resolution loss through down- and up-sampling (factor $1.5$--$4$), gamma ($[0.4, 1.8]$) and linear gain/bias shifts (gain $[0.85, 1.2]$, bias $[-0.1, 0.1]$), and a smooth multiplicative bias field (strength $[0.1, 0.4]$). Corruption is confined to the brain, leaving the skull-stripped background exactly at zero, and for 10\% of samples $x_0$ is replaced by pure noise so that the prior also learns to generate from scratch.

The second stage, \emph{cross-field finetuning} (Algorithm~\ref{alg:finetune}), learns the directed source-to-target mapping from the paired travelling cohort, iterating over all ordered field pairs $s \to \tau$ with $s \neq \tau$. Here the bridge runs between two real acquisitions of the same subject, the source slice as $x_0$ and the target as $x_1$. The source-field embedding $\mathrm{emb}_s(s)$ is introduced and trained at this stage, while the remaining weights are warm-started from the pretrained prior.

The third stage, \emph{adversarial refinement} (Algorithm~\ref{alg:refine}), initialises the generator from the finetuned model. To produce a slice it integrates the learned velocity outward from the source with Euler steps to $t=1$, yielding $\hat{x}_1$; this rollout is never conditioned on the target and therefore matches the inference procedure closely. Since Heun's method requires two network calls per step, in this training stage we use the simpler Euler method. The generator's gradient is backpropagated through the whole integration trajectory, so the adversarial signal from the discriminator refines every step of the transport rather than the endpoint alone.


\begin{figure}[hbt!]
\centering
\begin{minipage}[t]{0.49\textwidth}
\begin{algorithm}[H]
\footnotesize
\caption{Degradation-bridge pretraining (degraded target $\to$ target)}
\label{alg:pretrain}
\begin{algorithmic}
\State \textbf{Input:} unpaired slices; degradation menu $\mathcal{D}$; network $v_\theta$; EMA $\bar\theta$
\For{each minibatch $(x_1, \tau, m)$}
  \State $x_0 \gets m \odot \mathcal{D}(x_1)$
  \State $t \sim \mathcal{U}[0,1]$
  \State $x_t \gets (1-t)\,x_0 + t\,x_1$
  \State $v \gets v_\theta(x_t, t,\, \emptyset,\, \tau)$
  \State $\mathcal{L}_\mathrm{CFM} \gets \| m \odot (v - (x_1 - x_0)) \|^2 / \textstyle\sum m$
  \State $\theta \gets \mathrm{AdamW}(\theta, \nabla_\theta \mathcal{L}_\mathrm{CFM})$
  \State $\bar\theta \gets \mathrm{EMA}(\bar\theta, \theta)$
\EndFor
\State \Return $\bar\theta$
\end{algorithmic}
\end{algorithm}
\end{minipage}\hfill
\begin{minipage}[t]{0.49\textwidth}
\begin{algorithm}[H]
\footnotesize
\caption{Cross-field finetuning (source $\to$ target)}
\label{alg:finetune}
\begin{algorithmic}
\State \textbf{Input:} paired slices, all directed pairs
\State Initialise $\theta$ from pretraining EMA $\bar\theta$
\For{each minibatch $(x_0, x_1, s, \tau)$}
  \State $t \sim \mathcal{U}[0,1]$
  \State $x_t \gets (1-t)\,x_0 + t\,x_1$
  \State $v \gets v_\theta(x_t, t,\, s,\, \tau)$
  \State $\mathcal{L}_\mathrm{CFM} \gets \| v - (x_1 - x_0) \|^2$
  \State $\theta \gets \mathrm{AdamW}(\theta, \nabla_\theta \mathcal{L}_\mathrm{CFM})$
  \State $\bar\theta \gets \mathrm{EMA}(\bar\theta, \theta)$
\EndFor
\State \Return $\bar\theta$
\end{algorithmic}
\end{algorithm}
\end{minipage}
\end{figure}

\begin{algorithm}[hbt!]
\caption{Adversarial refinement}
\label{alg:refine}
\begin{algorithmic}
\State \textbf{Input:} paired slices; discriminator $D$; rollout steps $N_r$; weights $\lambda_\mathrm{adv}, \lambda_\mathrm{fm}$
\State Initialise $\theta$ from finetuning EMA $\bar\theta$; initialise discriminator $D_\phi$ with parameters $\phi$
\For{each minibatch $(x_0, x_1, s, \tau)$}
  \State $\hat{x}_1 \gets x_0$ \Comment{leak-free rollout: never sees $x_1$}
  \For{$i = 0$ \textbf{to} $N_r - 1$}
    \State $\hat{x}_1 \gets \hat{x}_1 + \tfrac{1}{N_r}\, v_\theta\!\big(\hat{x}_1, \tfrac{i}{N_r}, s, \tau\big)$ \Comment{$N_r$ Euler steps}
  \EndFor
  \State $\mathcal{L}_D \gets \mathrm{hinge}\big(D_\phi(x_1, x_0, s, \tau),\; D_\phi(\hat{x}_1, x_0, s, \tau)\big)$
  \State $\phi \gets \mathrm{AdamW}(\phi, \nabla_\phi \mathcal{L}_D)$
  \State $t \sim \mathcal{U}[0,1]$;\quad $x_t \gets (1-t)\,x_0 + t\,x_1$
  \State $\mathcal{L}_\mathrm{CFM} \gets \big\| v_\theta(x_t, t, s, \tau) - (x_1 - x_0) \big\|^2$ \Comment{normal training step like Alg. 2}
  \State $\mathcal{L}_\mathrm{adv} \gets -\,\mathrm{mean}\, D_\phi(\hat{x}_1, x_0, s, \tau)$
  \State $\mathcal{L}_\mathrm{fm} \gets \big\| D_\mathrm{feat}(\hat{x}_1) - D_\mathrm{feat}(x_1) \big\|_1$ \Comment{feature matching}
  \State $\mathcal{L}_G \gets \lambda_\mathrm{CFM}\mathcal{L}_\mathrm{CFM} + \lambda_\mathrm{adv}\,\mathcal{L}_\mathrm{adv} + \lambda_\mathrm{fm}\,\mathcal{L}_\mathrm{fm}$
  \State $\theta \gets \mathrm{AdamW}(\theta, \nabla_\theta \mathcal{L}_G)$;\quad $\bar\theta \gets \mathrm{EMA}(\bar\theta, \theta)$
\EndFor
\State \Return $\bar\theta$
\end{algorithmic}
\end{algorithm}

\section{Experiments and Results}
\label{sec:results}

\subsection{Implementation details}

\paragraph{Architecture:} The velocity network is a 2D U-Net \cite{unet} (6.3M parameters) with four resolution levels ([32, 64, 128, 128] channels) and two residual blocks per level \cite{resnet}. The adversarial discriminator is a two-scale conditional PatchGAN \cite{patchgan}; each scale is a three layer convolutional network (16 base channels) with a $34\times34$ receptive field.

\paragraph{Data handling:} We work in 2D, treating 3D volumes as stacks of 2D axial slices. A slice is kept for training only if its intensity standard deviation exceeds a threshold, which discards the empty slices outside the brain. Each slice is min--max normalised to $[0,1]$ and rescaled to $[-1,1]$ for the network. All computations happen in $[-1,1]$; outputs are mapped back to $[0,1]$ only when written out.

\paragraph{Optimisation:} Every stage uses AdamW \cite{adamw} with bf16 mixed precision and TF32 matmuls, and keeps an exponential moving average of the weights (decay 0.999). Pretraining runs on the unpaired cohort at a learning rate of $3\times10^{-4}$ with batch size 8 for 200{,}000 steps. Cross-field finetuning then runs on the paired cohort for 100{,}000 steps, dropping the learning rate to $5\times10^{-5}$ at the same batch size. Adversarial refinement continues for a further 50{,}000 steps at batch size 8: the generator (flow U-Net) learns at $1\times10^{-5}$ while the discriminator uses $3\times10^{-4}$, and the discriminator is trained alone for a 5{,}000-step warm-up before generator updates begin. The generator loss combines the velocity regression (weight 1), the adversarial term (weight $10^{-4}$, linearly ramped from zero over the first 2{,}000 generator steps), and the feature-matching term (weight 10). All experiments ran on a single NVIDIA RTX 6000 GPU (48\,GB) in PyTorch \cite{pytorch}.

\paragraph{Baselines:} We compare against two baselines that share our U-Net, the FiLM field conditioning, the data splits, and the two-stage training schedule, so that only the training objective differs. The first is a residual supervised regression trained with an L1 loss through the same degradation-bridge pretraining and paired finetuning; it translates in a single forward pass. The second is an $\epsilon$-prediction diffusion model with a cosine noise schedule, pretrained as a denoising prior on the unpaired cohort and finetuned to denoise a noised source towards its target; at inference the source is noised to $t=0.5$ and denoised back in 50 deterministic DDIM steps.

\subsection{Results and ablation}
\begin{figure}[t!b]
    \centering
    \includegraphics[width=\textwidth]{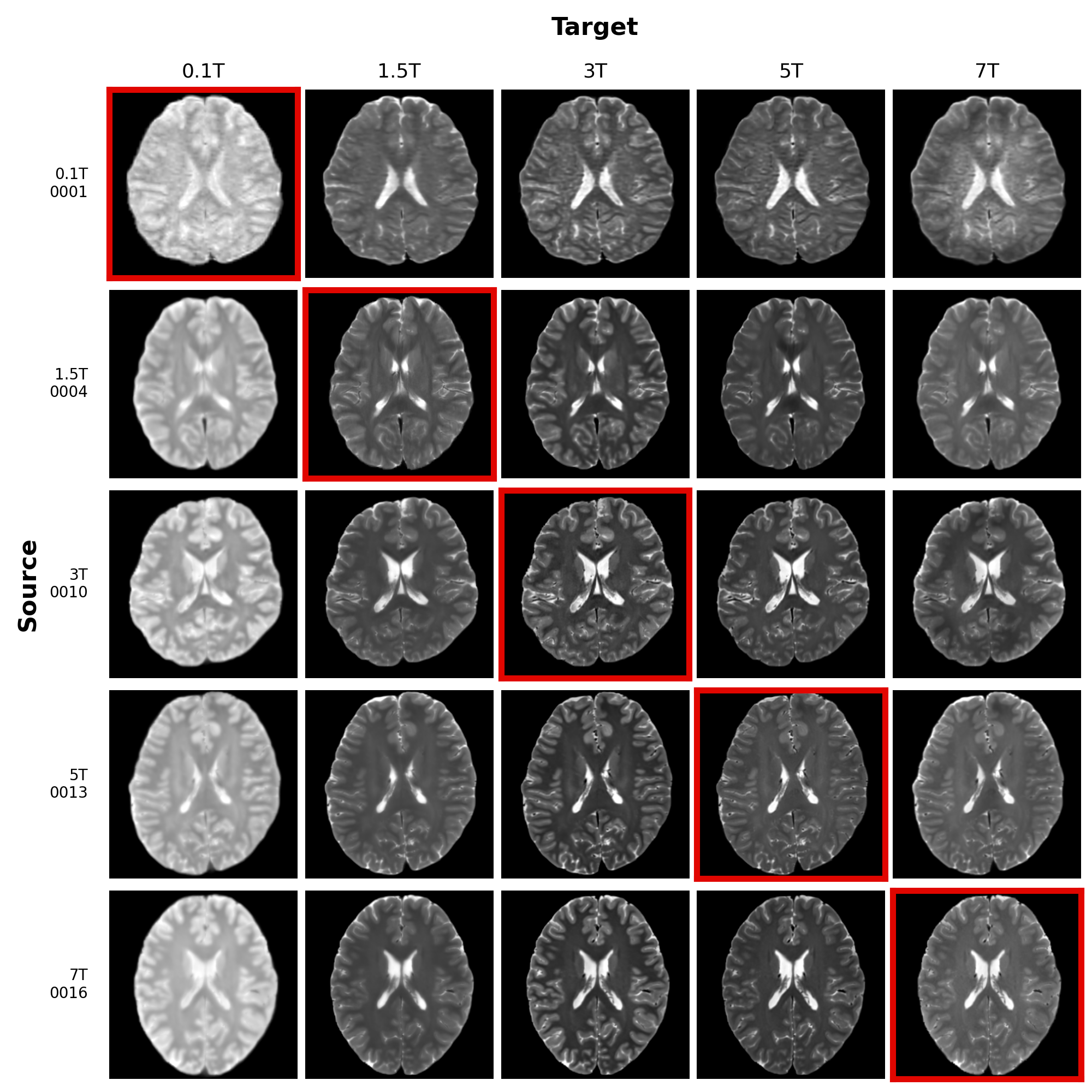}
    \caption{A series of example outputs from Task~3. The y-axis is the source field and the x-axis the target field. Scans with the red border are the ground truths for each target field, so every other scan is a prediction by our method.}
    \label{fig:res}
\end{figure}

Table~\ref{tab:results} reports the translation quality on the challenge evaluation set, scored by the challenge platform and averaged over all 20 directed field pairs and the three contrasts; each directed pair is evaluated on three held-out test subjects, a different trio per pair, and the platform reports aggregate metrics only. Per-contrast scores stay within 0.008 SSIM and 0.003 LPIPS of the average, so we report the contrast average throughout. As references we include an identity floor that copies the source unmodified, and the two baselines described above. Figure~\ref{fig:res} shows a representative T2w translation across the field transition. The identity floor shows how far the registered pairing alone carries: copying the source already gives 0.836 SSIM. Three comparisons stand out in Table~\ref{tab:results}. First, the baselines: the direct regression comes close on SSIM (0.906 vs.\ 0.909) but lags on perceptual quality (LPIPS 0.110 vs.\ 0.089) and reconstruction error (nRMSE 0.246 vs.\ 0.227), and the diffusion baseline trails on every metric. Since both baselines share our architecture, conditioning, data, and staged training recipe, this gap isolates the contribution of conditional flow matching itself. Second, the stage progression: the pretrained restoration prior alone, which never saw a field pair, already reaches 0.837 SSIM; paired finetuning adds 0.049, and adversarial refinement a further 0.024, with matching gains in LPIPS and nRMSE. Integrated in a single Heun step the refined model collapses to 0.817, because refinement trains the velocity field through a multi-step rollout and so calibrates it for multi-step integration; five steps recover 0.909 and ten bring no further gain.

\begin{table}[tb]
\centering
\caption{Translation quality on the challenge evaluation set, averaged over the three contrasts and all 20 directed field pairs. Top: identity floor and baselines sharing our architecture. Middle: our model after each training stage. Bottom: the final model integrated with 1, 5, and 10 Heun steps. The submitted configuration is in bold.}
\label{tab:results}
\begin{tabular}{llccc}
\toprule
Method & Sampler & SSIM $\uparrow$ & LPIPS $\downarrow$ & nRMSE $\downarrow$ \\
\midrule
Identity (copy source)    & ---            & 0.836 & 0.157 & 0.522 \\
Regression (same U-Net)   & 1 forward pass & 0.906 & 0.110 & 0.246 \\
Diffusion (same U-Net)    & 50 DDIM steps  & 0.896 & 0.121 & 0.249 \\
\midrule
Ours: pretrained only     & 5 Heun steps   & 0.837 & 0.147 & 0.468 \\
Ours: + finetuning        & 5 Heun steps   & 0.885 & 0.103 & 0.268 \\
Ours: + adv.\ refinement  & 1 Heun step    & 0.817 & 0.171 & 0.382 \\
Ours: + adv.\ refinement  & 5 Heun steps   & \textbf{0.909} & \textbf{0.089} & \textbf{0.227} \\
Ours: + adv.\ refinement  & 10 Heun steps  & 0.909 & 0.089 & 0.229 \\
\bottomrule
\end{tabular}
\end{table}

\section{Discussion and Conclusion}
We presented a conditional flow matching approach to cross-field brain-MRI harmonisation for the MRIxFields2026 challenge, addressing the any-to-any Task~3 with a single 6.3M model that bridges a registered source slice to its target under a conditioned velocity field, spanning all 20 directed field pairs across three contrasts. The design rests on a three-stage recipe: a degradation bridge distills the unpaired retrospective cohort into a restoration prior; cross-field finetuning then learns the directed mapping; and adversarial refinement sharpens the output while a velocity-regression anchor preserves fidelity.
Finetuned on only three paired travelling subjects, the model reaches a mean SSIM of 0.909, LPIPS of 0.089 and nRMSE of 0.227 over every directed pair and contrast, with a single set of parameters. It outperforms a direct regression baseline and a noised-source diffusion baseline sharing its architecture, conditioning, data, and recipe, tying the gain to the flow matching objective rather than the recipe or network. Distilling a restoration prior from the abundant \emph{unpaired} cohort frees the scarce paired data to learn the directed mapping rather than the underlying image prior from scratch, and five Heun steps match ten, keeping inference cheap.
The main limitation is that the model operates slice by slice in 2D and so does not enforce through-plane coherence. The adversarial refinement could in principle hallucinate texture; the faithfulness anchor keeps the reconstruction error improving under refinement, but an anatomy-focused validation remains future work. The clearest routes forward are a 2.5D or 3D formulation to recover that coherence, a broader set of simulated degradations that better matches real low-field physics, and a larger paired cohort as more travelling-subject data become available.
\section{Acknowledgments}
This publication is part of the project AI4AI with file number P22.017 of the research program 
Perspectief which is (partly) financed by the Dutch Research Council (NWO) under the grant TTW Perspectief 2022-2023. This work was performed using the SHARK HPC cluster provided by the LUMC.

\newpage
\bibliographystyle{splncs04}
\bibliography{refs}

\end{document}